\documentclass[conference]{IEEEtran}

\usepackage{times}
\usepackage{epsfig}
\usepackage{graphicx}
\usepackage{amsmath}
\usepackage{amssymb}
\ifCLASSINFOpdf
\else
\fi
\begin{document}

\title{S-TRIGGER: Continual State Representation Learning via Self-Triggered Generative Replay}

\author{\IEEEauthorblockN{Hugo Caselles-Dupr\'e\IEEEauthorrefmark{1},
Michael Garcia-Ortiz\IEEEauthorrefmark{2},
David Filliat\IEEEauthorrefmark{1}}
\IEEEauthorblockA{\IEEEauthorrefmark{1}U2IS, ENSTA Paris, Institut Polytechnique de Paris \& INRIA Flowers}
\IEEEauthorblockA{\IEEEauthorrefmark{2}CitAI, SMCSE, City University of London}}

\maketitle

\begin{abstract}

We consider the problem of building a state representation model for control, in a continual learning setting. As the environment changes, the aim is to efficiently compress the sensory state information without losing past knowledge, and then use Reinforcement Learning on the resulting features for efficient policy learning. 
To this end, we propose S-TRIGGER, a general method for Continual State Representation Learning applicable to Variational Auto-Encoders and its many variants. 
The method is based on Generative Replay, i.e. the use of generated samples to maintain past knowledge. It comes along with a statistically sound method for environment change detection, which self-triggers the Generative Replay. 
Our experiments on VAEs show that S-TRIGGER learns state representations that allows fast and high-performing Reinforcement Learning, while avoiding catastrophic forgetting. The resulting system has a bounded size and is capable of autonomously learning new information without using past data.

\end{abstract}

\section{Introduction}

The long term goal of creating agents capable of learning during extended periods of time, without constant supervision, will require the capability of continually learning about their environment.
Indeed, learning about the world is inherently a progressive/incremental task, since the data distribution experienced by the agent can change both spatially (e.g. countries) and temporally (e.g. seasons).
This involves building a model of an agent's surroundings, with visuals features, and continually updating this model as its life progresses and the environment evolves, without forgetting. 

\begin{figure}[h!]
    \centering
    \includegraphics[scale=0.25]{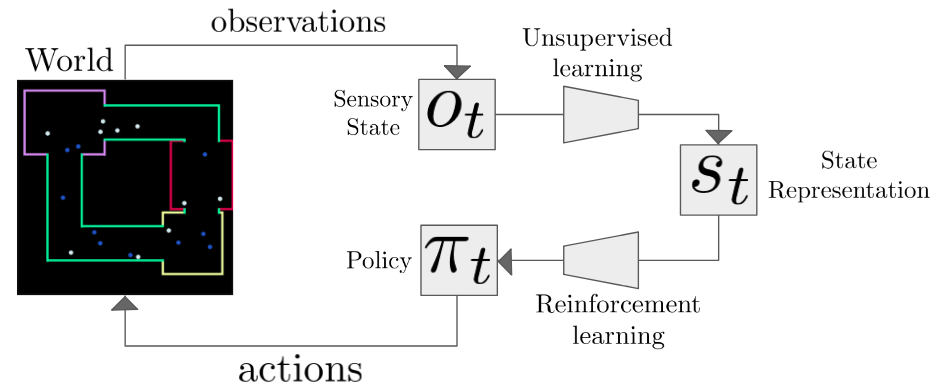}
    \caption{The State Representation Learning paradigm is decoupled in two phases: first learn a representation model for collected sensory states $o_t$, then feed the state representations  $s_t$ as input to a policy $\pi_t = \pi(s_t)$, which is optimized to solve a task using RL. Previous work \cite{raffin2019decoupling} 
    has shown this can be more efficient than using directly sensory states for RL.}
    \label{fig:srl}
\end{figure}

Recent advances in representation learning \cite{bengio2013representation} have brought successful tools for learning a model of the surroundings of the agent, a field known as State Representation Learning (SRL) \cite{lesort2018state}. Once trained, these representation models are used to learn policies using Reinforcement Learning (RL) more efficiently \cite{ha2018recurrent} than when using raw sensors. The process is presented in Fig.\ref{fig:srl}. However, previous approaches have mostly been limited to scenarios where the environment stays fixed. Since the models used are often neural networks trained using stochastic gradient descent (or any variant), they forget past knowledge when the training data distribution changes, an infamous problem called catastrophic forgetting \cite{french1999catastrophic}. In Continual SRL, the training data distribution changes as the environment evolves, which can happen in a discrete or continuous way. In our experiments, we consider sudden discrete changes of environment.

\begin{figure*}[t!]
    \centering
    \includegraphics[scale=0.4]{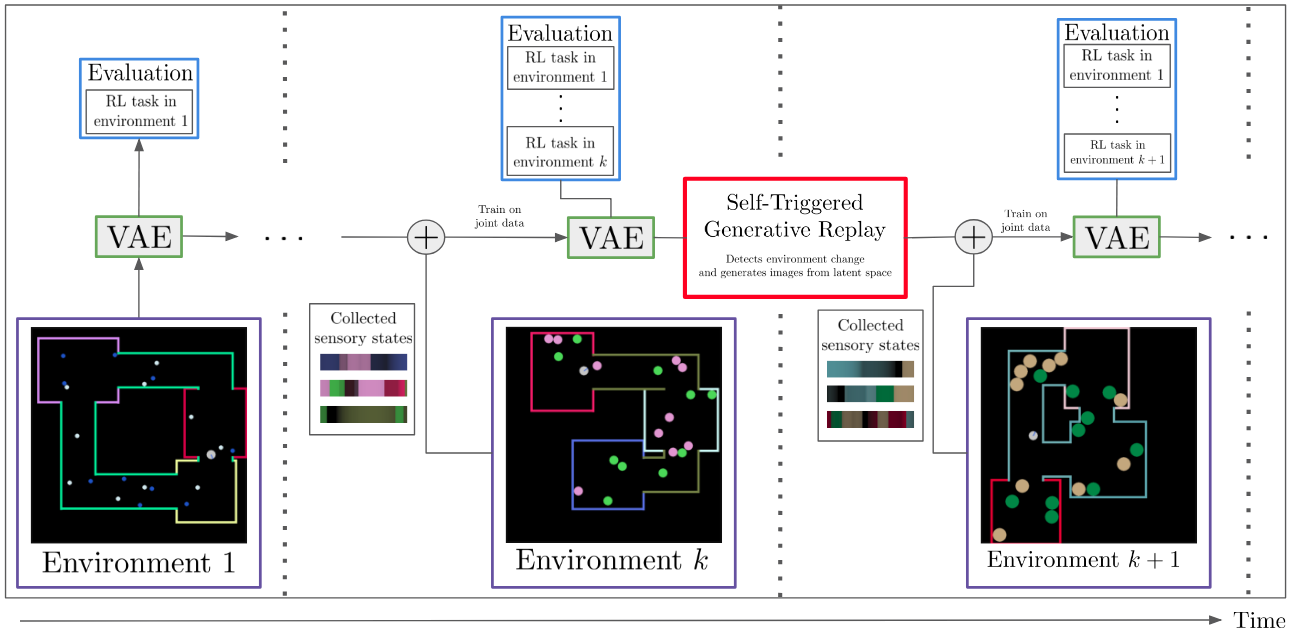}
    \caption{Overview of our proposed method S-TRIGGER for Continual State Representation Learning for RL. As the agent lives in environment 1, it collects sensory states (i.e. vision) to learn a state representation model (here, a VAE). We test this model by using it to solve, in an on-policy fashion, a RL task in environment 1. The agent is now moved to environment 2. Our method automatically detects this change, and use the VAE trained on environment 1 to generate images corresponding to environment 1, that are joint to collected sensory states of environment 2 to learn a state representation model for both environments. We test this state representation model by learning, in a on-policy fashion, to efficiently solve a task in environment 1 \textbf{and} 2. The procedure continues as the agent encounters new environments. Best viewed in color.}
    \label{fig:s-trigger}
\end{figure*}

We propose a general method for Continual SRL for agents behaving in a changing environment, termed Self-TRIGgered GEnerative Replay (S-TRIGGER for short), with the aim of learning representations useful for RL. Our approach is compatible with any type of State Representation model that is generative and reconstructs the input, which includes Varational Autoencoders (VAEs) and its many variants (e.g. $\beta$-VAE \cite{higgins2016beta}, ACN \cite{graves2018associative} and others) that are widely used for SRL. Previous work have shown that VAEs can learn continually using generated samples from previous tasks, a method called Generative Replay \cite{lesort2018generative}. S-TRIGGER uses Generative Replay to \textbf{remember information relative to previously encountered environments}. With the goal of an autonomous agent in mind, we complement our approach with a general method for \textbf{automatic detection of environment change}, based on the reconstruction error distribution of VAEs. This detection method allows the VAE to self-trigger the Generative Replay when a new environment is encountered, without the need for the user to specify environment changes. Finally, our approach respects important Continual Learning (CL) desiderata: no access to past data and bounded system size. Fig.\ref{fig:s-trigger} presents an overview of S-TRIGGER.


We focus on settings where the agent has only access to sensory states (here, vision) that contain partial information about the environment, similarly to an embodied agent. 
The agent should be able to use learned knowledge to facilitate policy learning with RL. Hence, we test our approach on a 2-D first-person simulator with coherent physics, in two scenarii where the environment changes over time. The first scenario is a proof of concept with 2 environments, with minimal change between the two. We then test the robustness of our approach in a more challenging setting: sequences of randomly generated mazes. We test whether the learned features provide efficient and high-performing RL training on a navigation task. We also measure the ability of our method to retain past knowledge, using reconstruction error and visualization. Our results show that our approach avoids catastrophic forgetting, provides features that enable efficient RL and detects environment changes almost perfectly. 

\section{Related work}

We first review how generative models are used for SRL, then previous work related to Continual SRL for RL.

\subsection{State Representation Learning using generative models}

Generative models are commonly used for SRL, as they can successfully learn compressed and useful representations of visual images. When trained on an agent's sensory inputs, they provide a vectorial representation of what the agent experience. They have recently shown promising results as State Representation models for RL \cite{ha2018recurrent, wayne2018unsupervised}. They can also generate data, i.e. generate sensory states that the agent could have experienced. 

Variational Auto-Encoders \cite{kingma2013auto} can be used as generative models by sampling a code $z$ and decoding it. They are often used for SRL \cite{van2016stable, ha2018recurrent}, along with other types of auto-encoders, because the reconstruction error provide an intuitive way of analyzing what the model has learned in the latent representation. VAEs have the additional advantage of being able to generate data, which can be used for Continual Learning, as described thereafter.

\subsection{Continual State Representation Learning for RL}

Generative models, e.g. VAEs, are widely used for SRL. Hence, one approach for Continual SRL is to apply a CL strategy developed for generative models. While most work in CL is focused on discriminative models, there is a recent interest in CL approaches for generative models \cite{achille2018life, nguyen2017variational}. Proposed methods often rely on using generated samples to avoid forgetting \cite{wu2018memory, lesort2018generative}. This technique, which we use in this paper, is termed Generative Replay. For instance it is used in the algorithm VASE \cite{achille2018life}, where the authors propose a representation model that can, similarly to S-TRIGGER, learn continually and detect changes in the data distribution. VASE dynamically allows spare representation capacity to account for newly acquired information. While their approach is promising, the learned features were not tested on a RL setting, for which our approach is specifically designed and tested. Previous work \cite{raffin2019decoupling} has demonstrated that learning state representation that perform well for policy learning is not straightforward. In our experiments, we demonstrate that S-TRIGGER can indeed continually learn features that allow fast and high-performing policy learning. 

Other types of continual learning strategies for generative models could be investigated. We decided to select Generative Replay because it respects important desiderata for CL: bounded system size, no access to previous data. We found no previous work on applying these CL approaches to State Representation Learning for RL.

Another line of work on Continual SRL for RL is presented in DARLA \cite{higgins2017darla}, which circumvent the problem of catastrophic forgetting by learning disentangled representations with a specific VAE architecture. 
Their approach learns factored features that are robust to minimal modifications of the factors of the environment, ours continually update features as environment changes are detected. They show that learning a policy in a source domain with their representation as input can lead to zero-shot transfer by using the same policy in a target domain (similar to the source domain). The idea of learning one state representation model that generalize to environments similar to previously encountered ones is important and compelling, but it is not clear how this model could handle heavy environment changes such as new objects appearing. In our experiments we shade light on the failure cases of this approach. However, both approaches are compatible so any progress on one side complements the other. 




\section{Continual State Representation Learning with Self-Triggered Generative Replay}

We now present our method for Continual State Representation Learning, S-TRIGGER, which is designed for settings where the environment changes discretely over time.

\begin{figure*}[h!]
    \centering
    \includegraphics[scale=0.5]{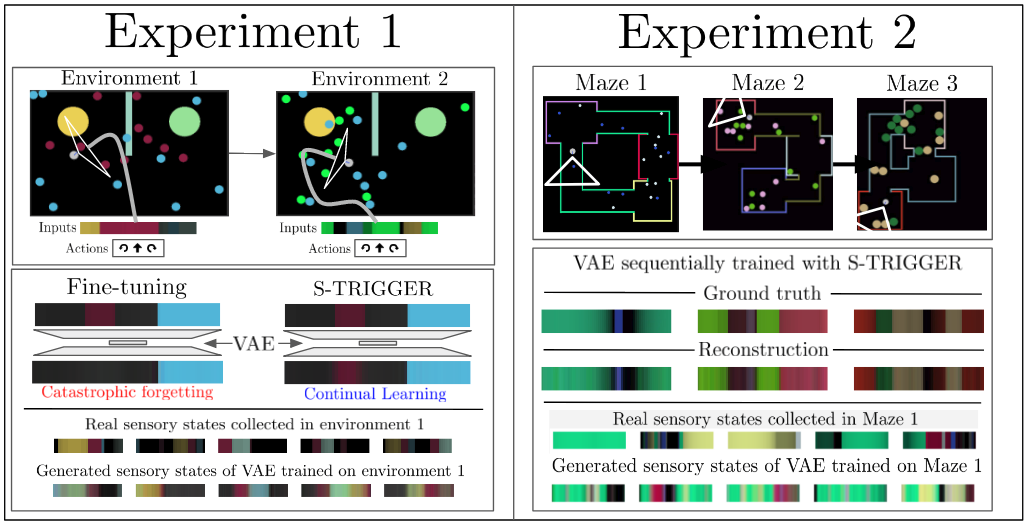}
    \caption{\textbf{Experiment 1}: \textit{Top}: The two environments considered in Experiment 1. \textit{Bottom}: Reconstruction comparison of sensory states from environment 1 between a VAE fine-tuned on environment 2 against a VAE trained with S-TRIGGER, followed by visualization of real and generated sensory states of a VAE trained in Environment 1. \textbf{Experiment 2}: \textit{Top}: One of the sequences of 3 randomly generated mazes considered in Experiment 2. \textit{Bottom}: Reconstruction of sensory states by a VAE sequentially trained with S-TRIGGER on the 3 mazes presented at the top, followed by visualization of real and generated sensory states of a VAE trained in Maze 1.}
    \label{fig:exp_1_exp_2}
\end{figure*}

\subsection{Learning continually with Generative Replay}
\label{sec:GR}
The considered problem is twofold: \textit{build a State Representation model} which can \textit{continually learn}, i.e. without forgetting. We use VAEs' encoding property 
to handle the State Representation and the generative ability of VAEs to generate states of previously seen environments to avoid forgetting, a technique known as Generative Replay \cite{shin2017continual}. In our case, we encode the sensory state received by the agent. 

Once an environment change is detected, we generate states using the latent space of the VAE trained on previous environments, and add those generated samples to states collected in the new environment. Then, we train the same VAE on the joint data. This procedure is illustrated in Fig.\ref{fig:s-trigger}. The environment changes are detected automatically, as described thereafter. 

This option does not re-use past data. Additionally, it is more scalable than a rehearsal method where samples from each previous environments would be maintained in memory, which amounts to scaling linearly in memory cost. Indeed, as new environments are encountered, we only need to maintain one VAE for all previously encountered environments, which makes Generative Replay a bounded system size solution in terms of memory cost. 

In theory, the training cost should scale linearly in number of environment changes. However, in our experiments, we use a fixed number $n$ of generated samples for each environment change, and we consider up to two environment changes (i.e. sequence of three  environments). $n$ is of the same order of magnitude as the number of collected samples in the environment. Hence, our method is bounded in terms of training cost for low number of environment changes. 

\subsection{Self-Trigger: Automatic environment change detection}
Since we aim at constructing an autonomous agent, we complement the proposed method with automatic detection of changes in the environment. Once an environment change is detected, Generative Replay is triggered, as described in Sec.\ref{sec:GR}, which is why we refer to our method as Self-TRIGgered GEnerative Replay (S-TRIGGER for short). 

The detection method is based on statistical hypothesis testing on VAE reconstruction error distribution. The intuition is: \textit{if the VAE reconstruction error distribution suddenly changes, then the environment must have changed.}. To formalize this intuition, we use Welch's $t$-test \cite{welch1947generalization} to test the hypothesis $\mathcal{H}_0$ that two sets
$x_1, x_2$ of randomly sampled VAE reconstruction errors have equal mean. We choose this test over the standard $t$-test because we do not have reasons to assume that the two samples variance are equal. The statistic $t$ is, under $\mathcal{H}_0$, distributed as a Student distribution with a number of degrees of freedom $\nu$ that can be approximated using the Welch-Satterthwaite equation.

with N being the number of samples (assumed equal for both samples), and $s_1, s_2$ being the empirical standard deviations of samples 1 and 2, respectively. The decision of the test is based on a chosen significance level $\alpha$ for the test, by using the $p$-value. If the $p$-value is below $\alpha$, it suggests that the observed data is sufficiently inconsistent with $\mathcal{H}_0$ that $\mathcal{H}_0$ may be rejected and thus an environment change is detected.



Using the $p$-value on a statistical test is preferable for decision over using a threshold-based method directly on the error distribution. The threshold method is based on an arbitrary scale which depends on the considered sequence of environments. On the contrary, the test is a more general approach based on statistical principles and thus agnostic to scales. The $p$-value provides an interpretable parameter for controlling the detection method's recall, which is not possible with the threshold method. Additionally, we choose to use VAE reconstruction error distribution over the actual state distribution because we are interested in changes of the environment that require the VAE to be updated. For instance, if we add an already existing obstacle to the environment, the state distribution shifts, while the VAE reconstruction error distribution does not change because the VAE needs not to be updated. The test is lightweight and can thus be continually performed.

\section{Experimental setting}
\label{sec:lesnoms_inputs}



\quad \textbf{Environments.} Our environments are developed in Flatland \cite{caselles2018flatland}. Flatland is a lightweight 2D environment for fast prototyping and testing of RL agents. It is of lower complexity compared to similar 3D platforms (e.g. DeepMind Lab or VizDoom), but emulates physical properties of the real world, such as continuity, partially-observable states with first-person view and coherent physics. It is suitable for prototyping CL approaches since many experiments are required and several enviroments are used, which can quickly become computationally prohibitive on 3D platforms. 


\quad \textbf{Methods.} For the State Representation model, we experiment with CCI-VAE \cite{burgess2018understanding}, the state-of-the-art VAE model for learning disentangled representations.

For our RL experiments, we learn to solve navigation tasks using features from the trained VAE. The policies are trained in an on-policy fashion. At each time-step $t$ we receive the sensory state input $o_t$, pass it through the SRL model (i.e. a VAE) to obtain the state representation $s_t$, and feed it to a policy $\pi(s_t)$ optimized with an RL algorithm, as described in Fig.\ref{fig:srl}. For the RL algorithm, we selected the Proximal Policy Optimization (PPO) algorithm \cite{schulman2017proximal} of SRL Toolbox \cite{raffin2018s}, which in our case employs an actor-critic network that both predicts the value function and a policy. We selected this method as it is a state-of-the-art policy gradient method, commonly used and robust to hyperparameter configurations. Policies are trained on-policy for $2M$ timesteps. Results are averaged over $5$ seeds, and we test the significance of our results with statistical hypothesis testing. To compare learning curves, we present smoothed (moving average) and normalized (by the maximum performance) mean reward curves.



\quad \textbf{Evaluation.} First, we evaluate whether the VAE successfully learned to reconstruct states, generate realistic states using sampling in the latent space and avoid catastrophic forgetting. We use visualization to measure performance. We also use the Mean Squared Error (MSE) to evaluate reconstruction quality and catastrophic forgetting over time as the environment changes.

Then, we evaluate the VAE ability to provide a state representation that enables efficient RL training, by measuring the performance of an RL agent using VAE features as input. The aim of SRL here is to have features that enable efficient behaviour learning hence the need to evaluate with RL. We use several baselines for comparison with S-TRIGGER (policy inputs are features of a VAE trained with our approach): Fine-tuning (policy inputs are features of a VAE sequentially fine-tuned), Source Only (policy inputs are features of a VAE trained only on first environment) and Upperbound (policy inputs are features of a VAE trained on data from all environment at once).

\section{Experiment 1: Proof of concept}

In this experiment we test S-TRIGGER on a continual learning setting with minimal changes between two environments. This experiment allows to easily visualize the failure mode of fine-tuning on this setting, and provides insights on how our method solves the presented issues. 

\begin{figure*}[h!]
    \centering
    \includegraphics[scale=0.45]{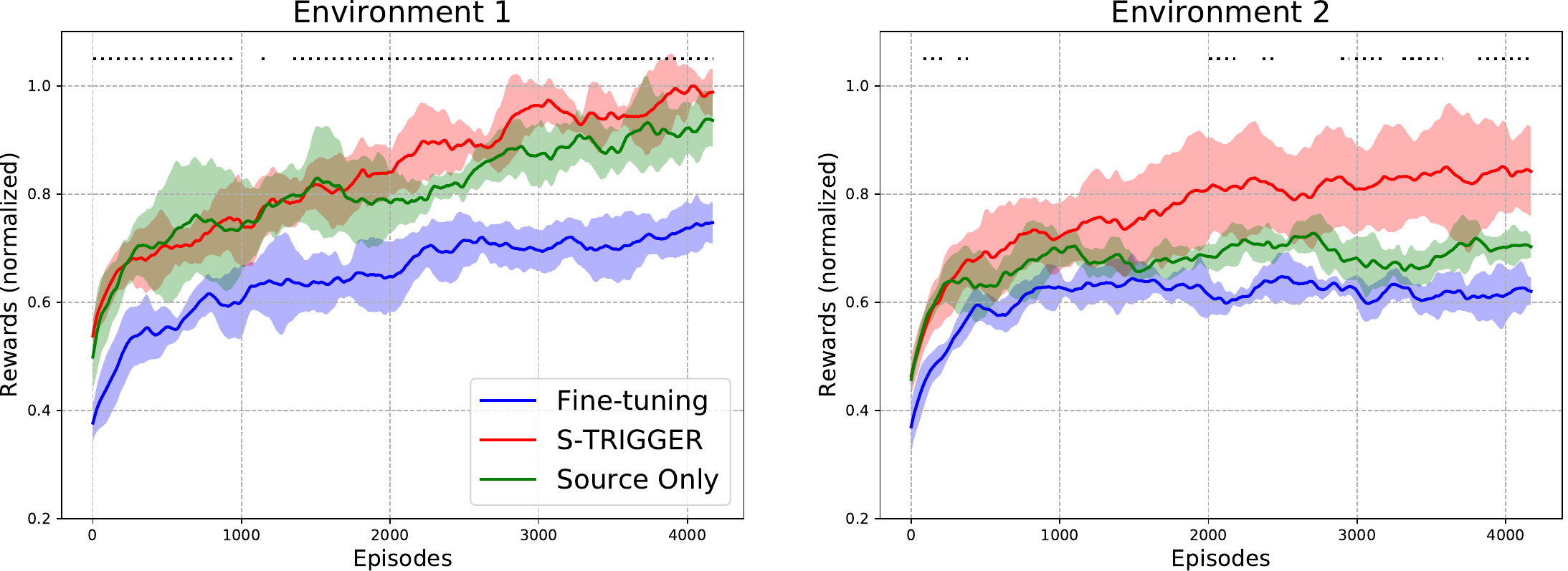}
    \caption{Smoothed normalized mean reward and standard error over 5 runs of RL evaluation using PPO with different inputs. Each method is described in Sec.\ref{sec:lesnoms_inputs}. 
    Dots indicates significance when testing against Fine-tuning with a Welch's
 t-test at level $\alpha = 0.05$.}
    \label{fig:rl_eval}
\end{figure*}

\subsection{Description of the environments}


We use two environments displayed in Fig.\ref{fig:exp_1_exp_2}. Most elements are identical between the two worlds: both are rooms of the same size with 3 fixed obstacles, 10 randomly placed round blue obstacles and 10 randomly placed round edible items. The only variation is the color of the edibles items. The environment changes once the representation model is finished training on the first environment. The navigation task consists in collecting as many edible items as possible in $500$ timesteps. The input is the vision of the agent, i.e. a 1-D image corresponding to what the agent sees in front of it. 

\subsection{Results}
\label{sec:results}



\quad \textbf{Reconstruction evaluation and detection of environment change.} We present in Fig.\ref{fig:exp_1_exp_2} a qualitative evaluation of Fine-tuning and S-TRIGGER methods on VAE. Models are sequentially trained on the first and then the second environment. Fine-tuning of VAE on the second environment leads to forgetting of the ability to reconstruct states from the first environment. On the contrary, S-TRIGGER successfully avoids this problem and the resulting VAE is able to properly reconstruct all elements of both environments, hence successful continual learning. Quantitatively the MSE of reconstruction over $500$ samples is similar for both methods on environment 2, whereas S-TRIGGER is one order of magnitude better than Fine-tuning on environment 1. This confirms our initial qualitative observations. We also present generated samples after the VAE is trained on the first environment in Fig.\ref{fig:exp_1_exp_2}. The VAE is able to generate realistic sensory states that are used by S-TRIGGER to remember past environments. 



Regarding environment change detection, our method tests whether a VAE trained on the first environment has a mean reconstruction error statistically different between states of environments 1 and 2. The reconstruction error mean is significantly higher on frames of environment 2, since the VAE has only been trained on frames from environment 1. We take advantage of this observation as we use a Welch's $t$-test to compare two batches of mean VAE reconstruction error, computed on randomly collected states from each environment. The null hypothesis is rejected if the $p$-value is greater than $0.01$. We repeat this experiment $5000$ times. The test is 100\% successful when it should detect an environment change, and 99.5\% successful when it should not detect a change. Any chosen critical $p$-value between $0.05$ and $0.0001$ provides similar results. 






\quad \textbf{RL evaluation.}  Learning curves and final performances are presented in Fig.\ref{fig:rl_eval}. We can observe catastrophic forgetting with Fine-tuning, as its performance is significantly inferior to other models on Environment 1. Indeed, this is expected since we previously showed that the model has forgotten how to reconstruct states of Environment 1. We also observe that VAE's features supports a form of zero-shot transfer. Indeed, using features of a VAE trained on the first environment only (see Source Only curve in Fig.\ref{fig:rl_eval}) to learn to solve Task $2$ is remarkably efficient. 

On Task 1, our method S-TRIGGER performs on par with a model trained only on Environment 1, which shows the absence of forgetting enabled by Generative Replay. On Task 2, the difference between models are less significant. However, S-TRIGGER still performs better or on par with Fine-Tuning. 
\begin{figure*}[!t]
    \centering
    \includegraphics[scale=0.4]{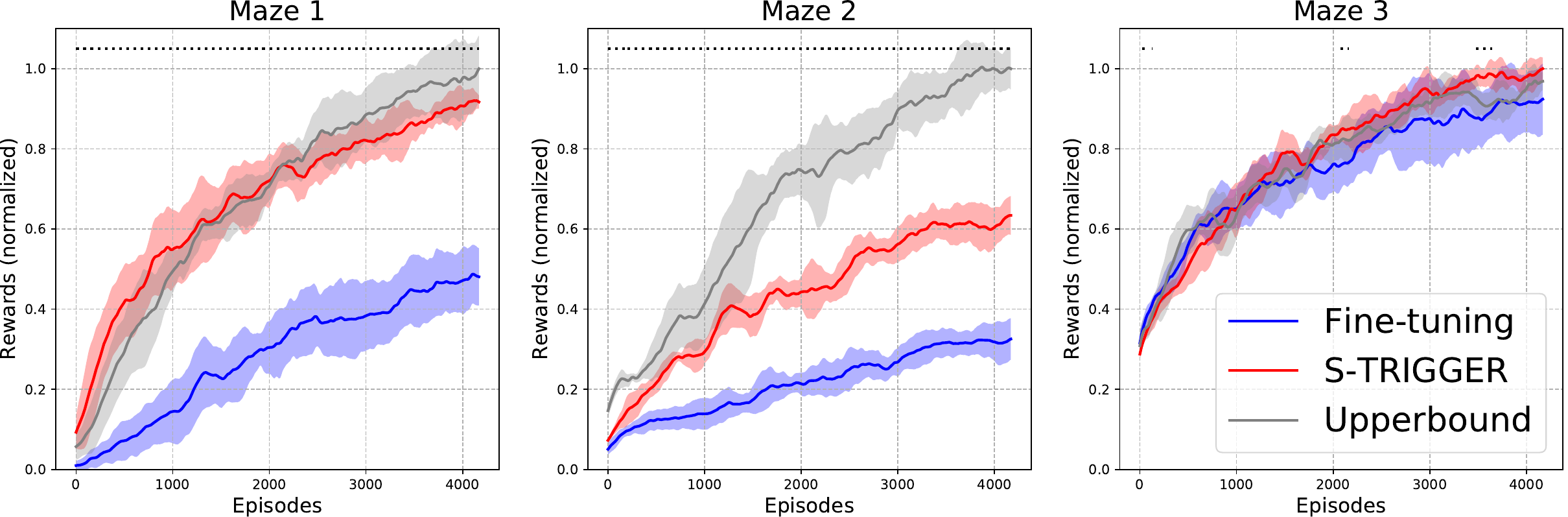}
    \caption{Randomly generated mazes: smoothed normalized mean reward and standard error over 5 runs of RL evaluation using PPO with VAE features as inputs. Each method is described in Sec.\ref{sec:lesnoms_inputs}. 
    Dots indicates significance when testing against Fine-tuning with a Welch's t-test at level $\alpha = 0.05$.}
    \label{fig:rl_eval2}
\end{figure*}

\section{Experiment 2: Robustness tests}

In this experiment, we test the robustness of S-TRIGGER on a scenario with sequences of randomly generated mazes. The setting is more challenging compared to Experiment 1 because of both the extended number of environments (3 sequential environments) and the heavier changes between them (changes of color, location, size of all elements). 

\subsection{Description of the environments}

We use sequences of three randomly generated mazes. The mazes are composed of rooms and corridors, and randomly placed edibles: fruits ($+10$ reward) and poisons ($-10$ reward). Fruits share the same color, different than poisons. In the sequence, rooms and corridors change location, size and color. The size of edibles gradually increase in the sequence, and their color changes as well. The environment changes once the representation model is finished training. In each of the three environments, the RL task is to learn how to collect fruits while avoiding poisons. The setting is illustrated in Fig.\ref{fig:exp_1_exp_2}.

\subsection{Results}
\label{sec:results2}

We first evaluate the reconstruction error and the environment change detection on a set of $100$ sequences of 3 randomly generated mazes. Then, we present the Reinforcement Learning evaluation on one sequence of 3 mazes.

\quad \textbf{Reconstruction evaluation and detection of environment change.} In this experiment, we compare S-TRIGGER and Fine-tuning by sequentially training a VAE on sequences of 3 randomly generated mazes, using randomly collected sensory states. We repeat the process $100$ times with different mazes each time. Contrary to Fine-tuning, S-TRIGGER does not forget how to reconstruct sensory states, and performs one order of magnitude better on previously encountered environments, i.e. Mazes 1 and 2. Qualitatively, we can observe how S-TRIGGER avoids catastrophic forgetting in Fig.\ref{fig:exp_1_exp_2}: after encountering Maze 3, the VAE reconstruction is satisfactory on all 3 mazes, hence the state information has been successfully compressed and this information has been kept over time. Also, the generation quality is still satisfactory, as seen in Fig.\ref{fig:exp_1_exp_2}.

Notice that the reconstruction is as satisfactory on Maze 1 as Maze 2 ($5.82\cdot 10^{-3}$ vs. $5.24\cdot 10^{-3}$), even though Maze 1 has been experienced by the agent more learning steps in the past than Maze 2. The VAE is still able to generate enough samples of Maze 1, and with a sufficient generation quality, to learn a satisfactory representation when Maze 3 is encountered. Also, since we obtained successful results on $100$ different sequences of 3 mazes, we can infer that S-TRIGGER is successful on any sequence of 3 randomly generated mazes. Our method is thus robust to the considered continual learning scenario.

As for the environment change detection method of S-TRIGGER, we test it on the 100 sequences, by artificially creating 400 transitions between environments per sequence (200 with a change, 200 without a change), which sums up to 4000 different transitions. The method performs quasi-perfectly well, with at least $99\%$ precision and recall. Considering the fact that it can be ran continuously for a low computational cost, the method is well suited for the considered setting.

\quad \textbf{RL evaluation.} For experiments in this section, we randomly selected one sequence on 3 randomly generated mazes, and kept it fixed for all experiments. VAE and policies are all trained on this same sequence of mazes, presented in Fig.\ref{fig:exp_1_exp_2}. We compare performance of policies trained on-policy with the RL algorithm PPO. Policies are trained using VAE features as inputs. VAEs are trained with three different strategies: Fine-tuning, S-TRIGGER (both sequentially trained) and Upperbound (jointly trained on data from the three mazes, not sequential). Learning curves and final performances are presented in Fig.\ref{fig:rl_eval2}. 

S-TRIGGER performs at least twice better than Fine-tuning on Maze 1 and 2, and performs optimally w.r.t Upperbound on Maze 1 and 3. Indeed, Upperbound has access to all data at once, which makes it an optimal solution w.r.t to using VAE features for policy learning in the considered Continual learning scenario. While S-TRIGGER cannot always reach the performance of this Upperbound (it does on Maze 1, but not on Maze 2), it clearly improves Continual SRL: the learned representation model allows fast and more efficient policy learning with a standard RL algorithm by limiting forgetting of previously seen environments.

Compared to Experiment 1, Fine-tuning performs worse on previously seen environments, due to the increased difference between encountered environments. As seen in the supplementary video, since the size of edibles increase for each new maze, a VAE sequentially trained with Fine-tuning until Maze 3 is not trained to see small objects in Maze 1, which makes the agent almost entirely \textit{blind} in Maze 1. On the contrary, S-TRIGGER allows the VAE to see objects in all mazes, which contributes to having faster and better-performing learned policies. Fine-tuning has one of the expected shortcomings of approaches like DARLA \cite{higgins2017darla}. When there is too much change between environments, learning a sufficiently general representation that can transfer to all environment is not feasible. The representation model needs to be updated when a new environment is encountered. 


\section{Conclusion}

We described S-TRIGGER, a Continual State Representation Learning method capable of learning as the environment changes discretely. 
Our method automatically detects changes and relies on using generated samples of previous environments, with a fixed-size system.
It learns a unique representation model that compresses information of each encountered environments, which can be used to train policies with RL, efficiently and with high-performance. 



\bibliographystyle{plain}
\bibliography{bibli} 

\end{document}